\documentclass{article}
\usepackage{spconf,amsmath,graphicx}
\usepackage{wrapfig}
\usepackage{graphicx}
\usepackage{amsmath}

\usepackage{amssymb}
\usepackage{mathtools}
\usepackage{amsthm}
\usepackage{booktabs}
\usepackage[switch]{lineno}
\usepackage{xcolor}
\usepackage{colortbl}
\usepackage{multirow}
\usepackage[ruled]{algorithm2e}

\title{Feature Projection Learning for Better Vision-Language Reasoning}

\name{Yi Zhang, Weicheng Lin, Liang-Jie Zhang\sthanks{Corresponding author.}}
\address{
\small{College of Computer Science and Software Engineering, Shenzhen University, China
}
}

\begin{document}
\ninept

\maketitle

\begin{abstract}
Vision-Language Pre-Trained models, notably CLIP, that utilize contrastive learning have proven highly adept at extracting generalizable visual features. To inherit the well-learned knowledge of VLP models for downstream tasks, several approaches aim to adapt them efficiently with limited supervision. 
However, these methods either suffer from limited performance, excessive learnable parameters, or extended training times, all of which hinder their effectiveness in adapting the CLIP model to downstream tasks. 
In this work, we propose a simple yet efficient and effective method called \textit{\textbf{F}eature \textbf{P}rojection \textbf{L}earning(FPL)} to address these problems. Specifically, we develop a projection model that projects class prototype features into the query image feature space and reconstructs the query image feature map. The negative average squared reconstruction error is used as the class score. In this way, we transform the classification problem into a feature projection problem.
The final output of this method is a combination of the prediction from the projection model and the original pre-trained CLIP. 
Comprehensive empirical evaluations confirm that FPL delivers superior accuracy, surpassing the current state-of-the-art methods by a substantial margin.

\end{abstract}

\begin{keywords}
Vision-Language Pre-Trained Models, CLIP, Feature Projection, Domain Generalization, Few-Shot Learning
\end{keywords}

\section{Introduction}
\label{sec:intro}

Foundational models have advanced significantly due to emerging studies on pre-trained Vision-Language Models (VLMs) such as CLIP \cite{radford2021learning}, which have demonstrated outstanding performance on open-vocabulary tasks. By harnessing visual semantics derived from massive collections of image-text pairs, these architectures exhibit remarkable proficiency across a wide range of downstream applications, often with zero-shot or few-shot learning \cite{radford2021learning,alayrac2022flamingo}.

Nevertheless, while zero-shot CLIP yields competitive results across general visual benchmarks, its reliance on pre-training often limits its ability to generalize effectively to specific, unencountered domains. Therefore, several works focus on fine-tuning these pre-trained VLMs for downstream tasks through designing learnable prompts derived from training instances. These  fine-tuning methods can be categorized into input-stage prompting such as CoOp \cite{zhou2022learning}, CoCoOp \cite{zhou2022conditional} and feature-stage fine-tuning such as TaskRes \cite{yu2023task}, CLIP-Adapter \cite{gao2021clip}, and Tip-Adapter \cite{zhang2022tip}. Specifically, CoOp \cite{zhou2022learning} pioneered the use of a learnable prompt to capture task-specific information. This approach was refined by CoCoOp \cite{zhou2022conditional}, which dynamically adjusts prompts on individual image instances. Furthermore, TaskRes \cite{yu2023task} introduces a decoupled task residual to ensure CLIP’s pre-existing knowledge remains intact. On the adapter front, CLIP-Adapter \cite{gao2021clip} employs an extra feature layer to enhance fine-tuning. Building on this, Tip-Adapter \cite{zhang2022tip} offers a training-free strategy by constructing a key-value cache from few-shot data; this cache can subsequently serve as a robust initialization for fine-tuning the keys (Tip-Adapter-F) to improve classification precision.

However, these methods either suffer from limited performance, excessive learnable parameters, or extended training times, all of which hinder their effectiveness in adapting the CLIP model to downstream tasks. For example, CoOp requires a substantial amount of training time but exhibits limited few-shot accuracy, While Tip-Adapter-F yields better performance, its large cache size introduces enormous learnable parameters. We then raise the question: \textit{Is it possible to improve the model's performance without introducing additional parameters or increasing the training time?}

To address the problems above, in this work, we propose a parameter-efficient method called Feature Projection Learning (FPL) to address these problems. 
We convert the classification problem into a projection problem. Specifically, given a set of images from the same class, we produce the associated feature maps and collect the component feature vectors across locations and images into a single pool of support features, we name it class prototype features.
For each query image, We then attempt to project prototype features of each class into the query image feature space, and  reconstruct the query image feature map as a weighted sum of class prototype features. The negative average squared  reconstruction error is used as the class score. Intuitively,  reconstructing images from the same class should be simpler because they share similar embeddings. Conversely, images from different classes will be more challenging to  reconstruct and may result in large  reconstruction errors.
Assessing the  reconstruction of the entire query image feature map enables us to retain the spatial details it contains. Additionally, permitting this  reconstruction to utilize feature vectors from any location in the support images intentionally eliminates irrelevant location-specific information.
Innovatively,  we approach feature map projection as a ridge regression task, which enables us to efficiently compute a solution using a straightforward closed-form approach, employing only a single learned, soft constraint. The final output of our proposed FPL is a combination of prediction from the projection model and original pre-trained CLIP.
Our extensive experimental results on few-shot classification and domain generalization demonstrate that our proposed FPL method can significantly improve CLIP's adaptation capability, and outperforms existing state-of-the-art methods, for example, by up to 5.1\% improvement on few-shot learning and by up to 4.3\% for domain generalization.

\section{Method}
\label{sec:method}

\begin{figure*}[!th]
\centering
\includegraphics[width=0.92\textwidth]{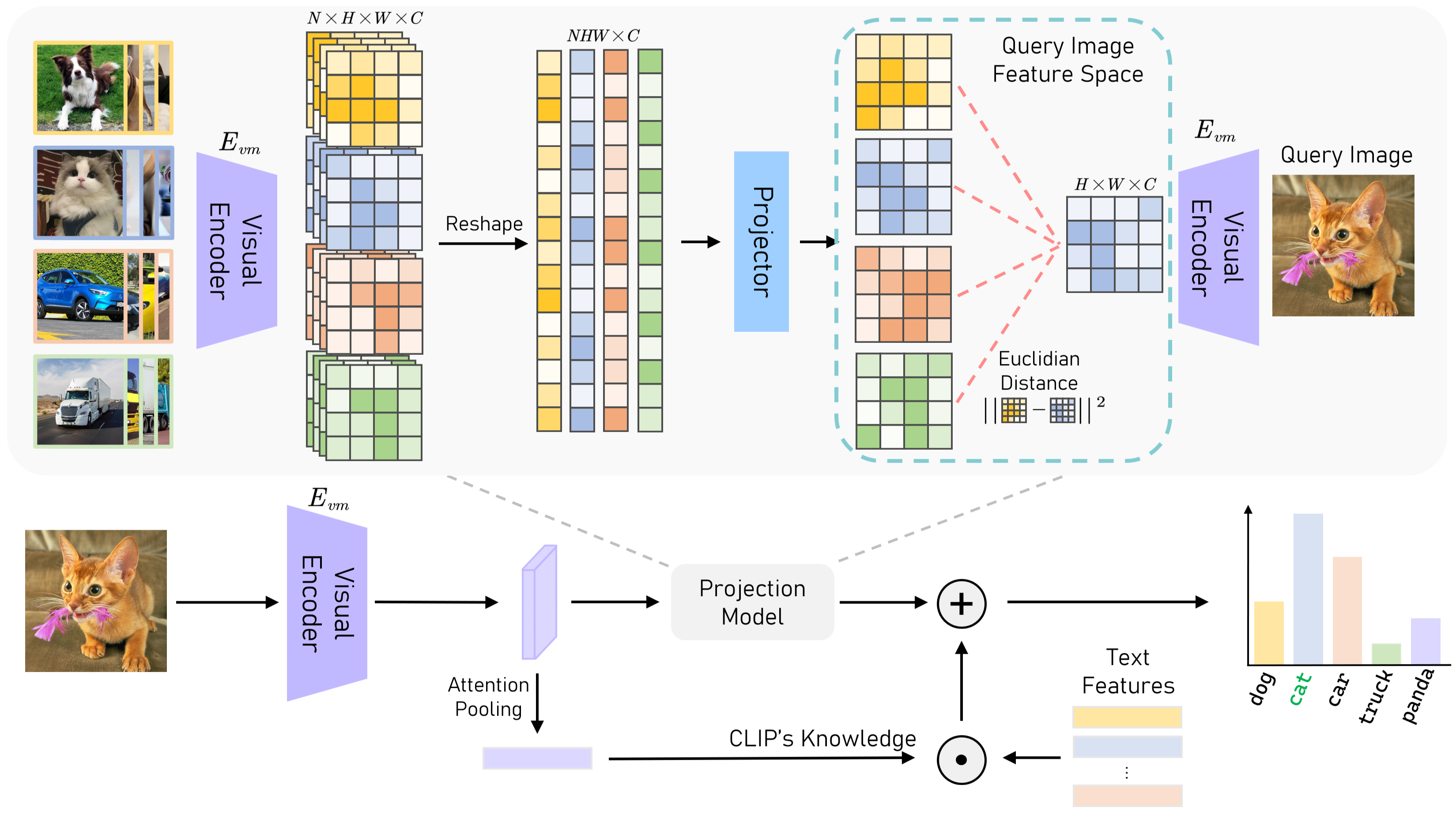}
\vspace{-8pt}
\caption{\textbf{Overview of our FPL method}. First, we generate the feature map of the query image by $E_{vm}$, for each class, we pool the features generated from the set of few-shot available images by $E_{vm}$ into a unified feature matrix, then we use projector to project these feature matrix into query image feature space and reconstruct the query image feature map. We calculate the negative squared Euclidean distance between the reconstructed query feature map and the real query feature map as the classification score. Combined with the prediction of the original CLIP, we got the final prediction. Note that $E_{vm}$ is the modified image encoder of CLIP that does not include the last attention pooling layer.}
\vspace{-5pt}
\label{fig:overview}
\end{figure*}

\subsection{A Revisit of CLIP}
CLIP \cite{radford2021learning} is a pioneering VLP model that consists of two parallel encoders for both textual and visual modalities. We denote these two encoders as $E_v$ and $E_t$ respectively. The two encoders of CLIP will map the text descriptions and images into the same embedding space. In the training stage, CLIP adopts a contrastive loss to align two modalities. In this work, we consider the problem of using CLIP for a $K$-classes image classification problem. For this specific setting, in the textual branch, we append a user-defined prompt such as $\sigma =$  ``a photo of ...'' with each class name $y_i$ in the set $Y=\{y_1, y_2, \cdots, y_K\}$ to get class-specific text descriptions. Using two encoders $E_v$ and $E_t$, we can generate the image feature $f_v$ and $K$ text feature $f_{t_i}$, which corresponds to class name $y_i$. Based on the cosine similarity of these features, we can make the prediction of the test image $X_{test}$ by
\begin{equation}
\label{eq-clip}
   p(y_i|X_{test})=\frac{\exp \left( \mathrm{sim}\left( f_{t_i},f_v \right) /\tau \right)}{\sum\nolimits_{j=1}^K{\exp \left( \mathrm{sim}\left( t_j,v \right) /\tau \right)}}, 
\end{equation}
where $\tau$ is the temperature hyper-parameter and $\mathrm{sim}(\cdot,\cdot)$ denotes the following cosine similarity:
\begin{equation}
\label{eq-sim}
   \mathrm{sim}\left( f_{t_i},f_v \right)=\frac{f_{t_i} \cdot f_v}{\Vert f_{t_i} \Vert  \Vert f_v \Vert}.
\end{equation}

\subsection{Prototype-to-Query Projection Model}
As shown in Figure \ref{fig:overview}, our model is built upon CLIP, using $E_t$ as the text encoder and $E_v$ as the image encoder. To begin with, we introduce the the architecture of visual encoder in detail. Take the ResNet encoder for example,  CLIP makes a small modification by appending an attention pooling layer to the conventional ResNet. We denote the encoder without attention pooling layer as $E_{vm}$. We observe that the output of $E_{vm}$ still retains sufficient spatial information and thereby can serve as a feature map. We denote the output as $\hat{M}\in \mathbb{R}^{H \times W \times C}$. where $H$, $W$, $C$ are the height, width, and the number of channels of the feature map respectively. Given $N$-shot $D$-class training samples, which means there are $N$ annotated images in each of $D$ classes. Our goal is  to predict a class label $y_q$ for a single query image $x_q$. 

We first generate the feature map of $x_q$ by $E_{vm}$, denoted as $M\in \mathbb{R}^{H \times W \times C}$. For each class $d\in D$, we pool the features extracted from the set of $N$ available images by $E_{vm}$  into a unified feature matrix denoted as $F_d\in \mathbb{R}^{NHW\times C}$. 
Our subsequent goal revolves around projecting $F_d$ into query image feature space, and reconstruct the matrix $M$ through a weighted combination of the rows within $F_d$. 

To achieve this, we define a projector 
which can project prototype image features into query image feature space, denoted as
\begin{align}
\label{eq:projector}
    Pr(F_d):\mathbb{R}^{NHW \times C} \rightarrow \mathbb{R}^{H \times W \times C} = \theta F_d.
\end{align}

We seek a matrix $\theta\in \mathbb{R}^{HW\times NHW}$ in such a way that the product $\theta F_d$ closely approximates $M$. Determining the optimal $\theta^*$ involves solving a linear least-squares problem:
\begin{align}
\label{eq:linprogram}
    \theta^* = \underset{\theta}{\text{arg min\ \ }} ||M-\theta F_d||^2 + \delta ||\theta||^2,
\end{align}
where $|| \cdot ||$ stands for the Frobenius norm, and $\delta$ serves as the parameter that controls the strength of the ridge regression penalty term. One of the primary advantages of the ridge regression formulation is its ability to yield a well-established closed-form solution for $\theta^*$ and the optimal  reconstruction $M_d^*$, as presented below:
\begin{align}
\label{q_origin}
    \theta^*  &= MF_d^T(F_dF_d^T+\delta I)^{-1},\\
    M_d^* &= \theta^* F_d = MF_d^T(F_dF_d^T+\delta I)^{-1}F_d.
\end{align}
For a specific class $d$, the scalar probability logit is determined by calculating the negative mean squared Euclidean distance between $M$ and $M_d^*$ across all locations of the feature map. It can be expressed as
\begin{align}
\label{q_mm}
    \langle M, M_d^* \rangle &= \frac{1}{HW}||M- M_d^*||^2.
\end{align}
The appropriate selection of the regularizer $\delta$ remains a non-trivial task. Instead of relying on heuristic methods, we opt for a more adaptive approach by enabling the network to learn $\delta$. This approach is particularly valuable as it gives the network the ability to ascertain an optimal level of regularization that promotes discriminative projection, rather than adhering rigidly to least-squares optimality.

To ensure non-negativity, we utilize $e^{\mu}$ to parameterize $\delta$, initialized to zero. Additionally, we introduce a trainable temperature factor denoted as $\epsilon$, inspired by the approach proposed in \cite{wertheimer2021few}. Consequently, the ultimate predicted probability is expressed as follows:
\begin{align}
\label{eq:q_full}
    \bar M_d =  \theta^* F_d &= MF_d^T(F_dF_d^T+e^{\mu}I)^{-1}F_d,\\
P_R(y_q=d|x_q) &= \frac{\exp{(-\epsilon\langle M,\bar M_d \rangle)} }
    {\sum_{d'\in D} \exp{(-\epsilon\langle M,\bar M_{d'} \rangle)} }.    
\end{align}
where $\mu$ and $\epsilon$ are learnable parameters. 

\subsection{Inference Fusion}
To process a specific class label $d \in D$, we insert it into a manual prompt template (e.g., "a photo of {class}"), which we represent as $\Pi_d$. $E_t$ is used to generate the corresponding text feature $f_t^d$, represent as $f_t^d = E_t(\Pi_d)$. First, we utilize $E_v$ to extract the feature $f_{v}$ from the image $x_q$. Given that both $f_{v}$ and $f_t$ are $L2$-normalized, for the original CLIP, according to Equal (\ref{eq-clip}), the prediction probability of $x_q$ belongs to class d is:
\begin{equation}
\label{eq:clip-logit}
 P_{CLIP}(y_q=d|x_q) = \frac{\exp{(sim\left( f_{v},f_t^d \right) /\tau} )}
    {\sum_{d'\in D} \exp{(sim\left( f_{v},f_t^{d'} \right) /\tau} )},   
\end{equation}
where $\tau$ is the learned temperature parameter of CLIP.

To enhance classification accuracy, we integrate the visual representation reconstruction model with the original CLIP inference. 
The final predicted probability for the input image $x_q$ is:
\begin{align}
\label{eq-LOGIT}
P_{total}(y_q=d|x_q) \!=\! P_{CLIP}(y_q\!=\!d|x_q) + \eta P_R(y_q\!=\!d|x_q),
\end{align}
where $\eta$ is a hyper-parameter to control the scaling of the residual connection. We use cross-entropy loss for classification, denoted as
\begin{equation}
    \mathcal{L}_{ce} = - \underset{\mu, \epsilon,}{\arg\min} \  \underset{(x,y) \in \mathcal{D}_{tr}}{\mathbb{E}} \sum_{k=1}^{\mathcal{Y}_{tr}} y_{k} \log(p(y_k|x)),
    \label{eq:lce}
\end{equation}
where $\mathcal{D}_{tr}$ is the few-shot training set, $\mathcal{Y}_{tr}$ are the class labels for training set. 
Generally, in our method, only $\mu$ and $\epsilon$ are needed to be updated by gradient descent. 

\textbf{Projection Orthogonality.}  To ensure that prototype features from different classes, when projected into the query image feature space, are distinct from each other, we utilize a projection orthogonality loss, which encourages the post-projection features of different classes to be dissimilar within the query image feature space:

\begin{footnotesize}
\begin{equation}
    \mathcal{L}_{po}= \frac{1}{D(D-1)}\sum_{i=1}^D \sum_{j=i+1}^D |< M_i^*, M_j^*>|,
\label{eq:8}
\end{equation}
\end{footnotesize}

\noindent where $<\cdot,\cdot>$ denotes the cosine similarity, $M_i^*$ represents the post-projected features in query image feature space from class $i$. Then the total training loss is:
\begin{equation}
    \mathcal{L}=\mathcal{L}_{ce} + \gamma \mathcal{L}_{po},
\label{eq:loss_all}
\end{equation}

\noindent where $\gamma$ is a hyper-parameter. We set $\gamma=0.1$ for all experiments.

\section{Experiment} 
\label{sec:experiment}

\subsection{Experiment Setup} \label{sec:setup}

To thoroughly validate the efficacy of our approach, we performed extensive experiments covering both few-shot image recognition and domain generalization. In the \textbf{few-shot image recognition} setting, we followed common evaluation standards using a collection of 11 widely used datasets. We examined performance on generic object classification (ImageNet \cite{recht2019imagenet}, Caltech101 \cite{fei2004learning}), fine-grained object classification (OxfordPets \cite{parkhi2012cats}, StandfordCars \cite{krause20133d}, Flowers102 \cite{nilsback2008automated}, Food-101 \cite{bossard2014food}, FGCV Aircraft \cite{maji2013fine}), texture classification (DTD \cite{cimpoi2014describing}), remote sensing recognition (EuroSAT \cite{helber2019eurosat}), scene recognition (SUN397 \cite{xiao2010sun}) and action recognition (UCF101 \cite{soomro2012ucf101}). 
Collectively, these datasets serve as a thorough standard for assessing few-shot capabilities. To validate \textbf{domain generalization}, we test the model's robustness under distribution shifts; specifically, we train using 16-shot ImageNet~\cite{deng2009imagenet} and evaluate performance across four ImageNet derivatives: ImageNet-V2~\cite{recht2019imagenet}, ImageNet-Sketch~\cite{wang2019learning}, ImageNet-A~\cite{hendrycks2021natural}, and ImageNet-R~\cite{hendrycks2021many}. These variant datasets are effective out-of-distribution data for ImageNet in previous research \cite{zhou2022conditional,shu2022tpt}.

\subsection{Implementation Details} \label{sec:implementation}

We build upon the CLIP architecture \cite{radford2021learning}, employing a ResNet-50 visual backbone alongside a Transformer-based text processor. We maintain the model's pre-trained capabilities by keeping the weights frozen throughout training. As outlined in Section \ref{sec:method}, the component $E_{vm}$ represents the modified image encoder of CLIP that does not include the last attention pooling layer, whose output is feature maps that will be used in the prototype-to-query projection model.
Across all benchmarks, we optimized the model using AdamW \cite{kingma2015adam} with an initial learning rate of $1\times 10^{-3}$, modulated by a cosine annealing scheduler. Due to the lightweight and parameter-efficient nature of our architecture, training was conducted on a single NVIDIA RTX 3090 GPU. We adhered to standard few-shot protocols, training on randomly sampled sets of 1, 2, 4, 8, and 16 instances per category and assessing performance on the complete test set.

\subsection{Performance Analysis} \label{sec:Performance Analysis}

\begin{figure*}[t]
\centering
\includegraphics[width=\textwidth]{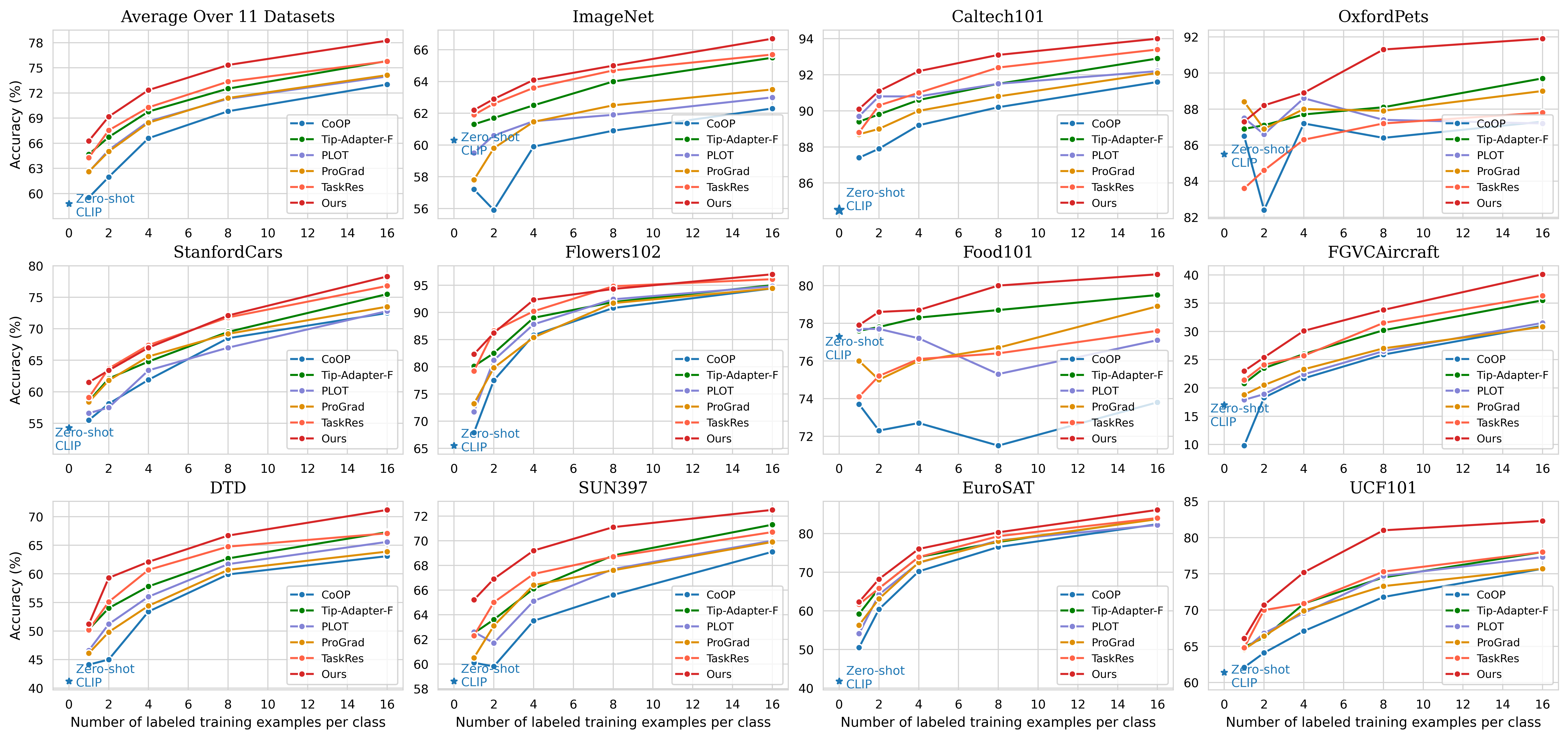}
\vspace{-20pt}
\caption{\textbf{Performance comparisons on few-shot learning on 11 datasets}. For each dataset, we report the accuracy on 1-/2-/4-/8-/16-shot settings. The top-left subfigure shows the average accuracy over all 11 datasets.}
\label{fig:fewshot_results}
\vspace{-5pt}
\end{figure*}

\subsubsection{Few-Shot Recognition}

As illustrated in Figure \ref{fig:fewshot_results}, we compare our approach against five baseline methods across the full suite of 11 datasets. The upper-left sub-figure shows the average accuracy, indicating that our method maintains the lead by exceeding other state-of-the-art techniques. Notably, compared to Tip-Adapter-F \cite{zhang2022tip}, our method achieves substantial performance improvements across all datasets tested. Especially, for difficult tasks such as fine-grained recognition, remote sensing recognition, action recognition, and texture classification, FPL outperforms Tip-Adapter-F by up to 5.1\% on UCF101, 2.2\% on EuroSAT, 4.5\% on FGCVAircraft, 4.2\% on DTD. It demonstrates the superior effectiveness in adapting CLIP to downstream tasks.  

\begin{table}[t]

\small
\begin{center}
\vspace{-15pt}
\caption{\textbf{Performance comparisons on domain generalization}. All experiments are conducted with ResNet-50 visual backbone. The best results are in \textbf{bold} and the second are \underline{underlined}.}
\vspace{-5pt}
\label{table:generalization}
\resizebox{\linewidth}{!}{
\begin{tabular}{lcccccc}
\toprule
\multirow{2}*{Method}  & Source & \multicolumn{5}{c}{Target} \\ \cmidrule(lr){2-2} \cmidrule(lr){3-7}  & ImageNet & -V2 & -Sketch & -A & -R  & Avg. \\
\midrule

Zero-Shot CLIP~\cite{radford2021learning}  &  60.33 &  53.27  & 35.44  & 21.65 &  56.00  & 41.59\\
CoOp~\cite{zhou2022learning}   & 63.33  & 55.40  & 34.67  & 23.06  & 56.60 &  42.43\\
CoCoOp~\cite{zhou2022conditional} &   62.81  & 55.72  & 34.48  & 23.32  & 57.74  & 42.82\\
ProGrad~\cite{zhu2022prompt} &   62.17  & 54.70  & 34.40  & 23.05  & 56.77  & 42.23\\
PLOT~\cite{chen2023plot} &   63.01  & 55.11  & 33.00  & 21.86  & 55.61  & 41.40\\
TPT~\cite{shu2022tpt} &   60.74 &  54.70  & 35.09  & 26.67  & \underline{59.11}  & 43.89\\

\rowcolor{gray!20}
\textbf{FPL (Ours)} &   \textbf{66.68}  & \textbf{57.66}  & \textbf{36.78}  & \textbf{30.93}  & \textbf{60.37}  & \textbf{46.46}\\
\bottomrule
\end{tabular}
}
\end{center}
\vspace{-15pt}
\end{table}

\begin{table}[t]
\centering
\vspace{-5pt}
\caption{\textbf{Efficiency comparisons on 16-shot ImageNet}. We report the results using a single NVIDIA RTX 3090 GPU.}
\vspace{2pt}
\label{table:efficiency}
\resizebox{\linewidth}{!}{
\begin{tabular}{l|cccccc}
\toprule
Method   & Epochs    & Training & GFLOPs  & Param. & Acc. \\
\midrule
CoOp & 200   & 15 h & $>$10 & 0.01M   & 62.95 \\
CLIP-Adapter & 200   & 50 min & 0.004 & 0.52M   & 63.59 \\
Tip-Adapter-F & 20   & 5 min & 0.030 & 16.3M   & 65.51 \\
\rowcolor{gray!20}
\textbf{FPL (Ours)} & \textbf{20}   & \textbf{1 min} & \textbf{0.001} & \textbf{0.001M}   & \textbf{66.46} \\
\bottomrule
\end{tabular}
}
\vspace{-10pt}
\end{table}

\subsubsection{Domain Generalization}

Table \ref{table:generalization} presents the performance metrics of our method alongside other state-of-the-art methods. For comparison fairness, we use the official figures from the original papers for all baselines. The analysis covers the source accuracy on ImageNet \cite{deng2009imagenet}, the target accuracy on various shifts (ImageNet-V2~\cite{recht2019imagenet}, ImageNet-Sketch~\cite{wang2019learning}, ImageNet-A~\cite{hendrycks2021natural}, ImageNet-R~\cite{hendrycks2021many}), and the overall average. The data clearly shows that our approach significantly outperforms all baselines across all metrics, underscoring remarkable robustness to distribution shifts.

\subsection{Efficiency Comparison}

To evaluate efficiency, we used a single NVIDIA RTX 3090 GPU. We compared our method with other state-of-the-art methods in terms of training epochs, training time, computational cost, and parameter count on 16-shot ImageNet. The comprehensive results are detailed in Table \ref{table:efficiency}.
With just 1 minute, 0.001GFLOPs, and 0.001M parameters for training, our proposed method achieves a remarkable accuracy of 66.68\% on 16-shot ImageNet. 
In comparison, the CoOp method needs about 15 hours of training and over 10 GFLOPs to achieve 62.26\% accuracy; the Tip-Adapter-F method needs 5 minutes of training and 0.03 GFLOPs to achieve  65.51\% accuracy.

\subsection{Ablation Study}

To verify the contribution of individual modules within our proposed method, we performed an ablation analysis using the ImageNet dataset \cite{deng2009imagenet}. The detailed results of the experiments are provided in Table \ref{table:ablation}, where the last row shows the accuracy of our complete FPL.
We've found that both ridge regression penalty and projection orthogonality loss have a significant impact on enhancing model performance, with ridge regression penalty having a relatively greater influence. projection orthogonality loss has a more pronounced effect in few-shot settings.

\begin{table}[t]
\vspace{-14pt}
\caption{\textbf{Effectiveness of different components in our FPL method}. \textit{w/o} PO represents models without projection orthogonality loss,  fixed $\delta$ denotes that we fix $\delta$ and set $\mu=0$.}
\vspace{3pt}
\label{table:ablation}
\centering
\resizebox{\linewidth}{!}{
\begin{tabular}{lccccc}
\toprule
Few-shot Setup & 1    & 2  & 4  & 8 & 16 \\ \midrule
\textbf{FPL(ours)} & \textbf{62.15} & \textbf{62.86} & \textbf{64.13} & \textbf{65.02} & \textbf{66.68} \\
\midrule
 \textit{w/o} PO         & 61.82 & 62.68 & 63.93 & 64.92 & 66.51 \\
fixed $\delta$       & 60.86 & 61.12 & 61.25 & 61.87 & 62.65 \\

\textit{w/o} PO + fixed $\delta$            & 60.66 & 60.93 & 61.02 & 61.77&62.56 \\
\bottomrule
\end{tabular}
}
\vspace{-7pt}
\end{table}

\section{Conclusion}
\label{sec:conclusion}
We introduce Feature Projection Learning (FPL), a novel method for adapting CLIP to few-shot classification and domain generalization by projecting class prototype features into query image feature space. Our proposed FPL is simple yet effective and efficient with introducing a minimal number of learnable parameters. We demonstrate our method achieves state-of-the-art performances on 11 few-shot classification datasets and 4 domain generalization datasets.

\clearpage
\bibliographystyle{IEEEbib}
\bibliography{strings,refs}

\end{document}